\pdfoutput=1
\PassOptionsToPackage{table}{xcolor}
\documentclass[11pt]{article}

\usepackage[final]{acl}

\usepackage{times}
\usepackage{latexsym}

\usepackage[T1]{fontenc}

\usepackage[utf8]{inputenc}

\usepackage{microtype}

\usepackage{inconsolata}

\usepackage{graphicx}

\usepackage{tcolorbox}
\usepackage{amssymb}
\usepackage{booktabs}
\usepackage{amsmath}

\definecolor{dodgerblue}{rgb}{0.12, 0.56, 1.0}
\definecolor{forestgreen}{rgb}{0.13, 0.55, 0.13}

\definecolor{economics}{rgb}{0.282, 0.471, 0.816}
\definecolor{eecs}{rgb}{0.933, 0.522, 0.29}
\definecolor{law}{rgb}{0.835, 0.733, 0.404}
\definecolor{math}{rgb}{0.51, 0.776, 0.886}
\definecolor{medicine}{rgb}{0.584, 0.424, 0.706}
\definecolor{natural}{rgb}{0.549, 0.38, 0.235}
\definecolor{politics}{rgb}{0.863, 0.494, 0.753}
\definecolor{psychology}{rgb}{0.475, 0.475, 0.475}

\newif\ifrev
  \revtrue
\ifrev

\else

\fi

\title{Designing Role Vectors to Improve LLM Inference Behaviour}

\author{
 \textbf{Daniele Potertì\textsuperscript{2}},
 \textbf{Andrea Seveso\textsuperscript{1,3}},
 \textbf{Fabio Mercorio\textsuperscript{1,3}}
\\
\\
  \textsuperscript{1}Dept of Statistics and Quantitative Methods, University of Milano-Bicocca, Italy,
 \\
 \textsuperscript{2}Dept of Economics, Management and Statistics, University of Milano-Bicocca, Italy,
 \\
 \textsuperscript{3}CRISP Research Centre \url{crispresearch.eu},  University of Milano-Bicocca, Italy
}

\begin{document}
\maketitle
\begin{abstract}
    The influence of personas on Large Language Models (LLMs) has been widely studied, yet their direct impact on performance remains uncertain. This work explores a novel approach to guiding LLM behaviour through role vectors, an alternative to persona-based prompting. We construct 29 role vectors derived from model activations and evaluate their impact on benchmark performance across multiple domains. Our analysis investigates whether these vectors can effectively steer models toward domain-specific expertise. We measure two key interventions: (i) activation addition, which reinforces role-specific directions, and (ii) directional ablation, which removes them. Results on well-established benchmarks indicate that role vectors do, in fact, influence model behaviour, improving task performance in relevant domains while marginally affecting unrelated tasks. This, in turn, suggests that manipulating internal model representations has a greater impact on outcomes than persona-based prompting.
\end{abstract}

\section{Introduction}
\label{sec:intro}

The development of persona or role-based chatbots has gained significant attention in the AI and NLP community due to their potential impact on business and societal applications~\cite{pataranutaporn2021ai}. The extent to which different personas influence Large Language Models' (LLMs) performance on objective tasks remains unclear. Recent attempts investigate whether incorporating personas into system prompts enhances model performance on objective tasks and explores potential factors influencing these effects. \cite{zheng2024helpful} conducted a large-scale analysis of the effect of personas in LLM prompting, examining the impact of domain alignment between personas and task-related questions, finding that persona-based prompting either has no effect or a slightly negative impact on model performance compared to a baseline setting.

We aim to investigate whether modifying the model’s internal mechanisms~\cite{li2024inference}, rather than a prompt-based approach, can lead to improved results. This forms the core objective of our current work, guided by the following research questions:
\textbf{RQ1}: Can we identify specific latent role directions within the activation space, derived from the model's internal mechanisms, that, when leveraged, lead to improved performance on objective tasks?
\textbf{RQ2}: Do the directions that enhance performance effectively impersonate the role of interest?
\textbf{RQ3}: If we eliminate these directions in the models, do their performances suffer as a consequence?

\begin{figure}[t]
    \includegraphics[width=1\linewidth]{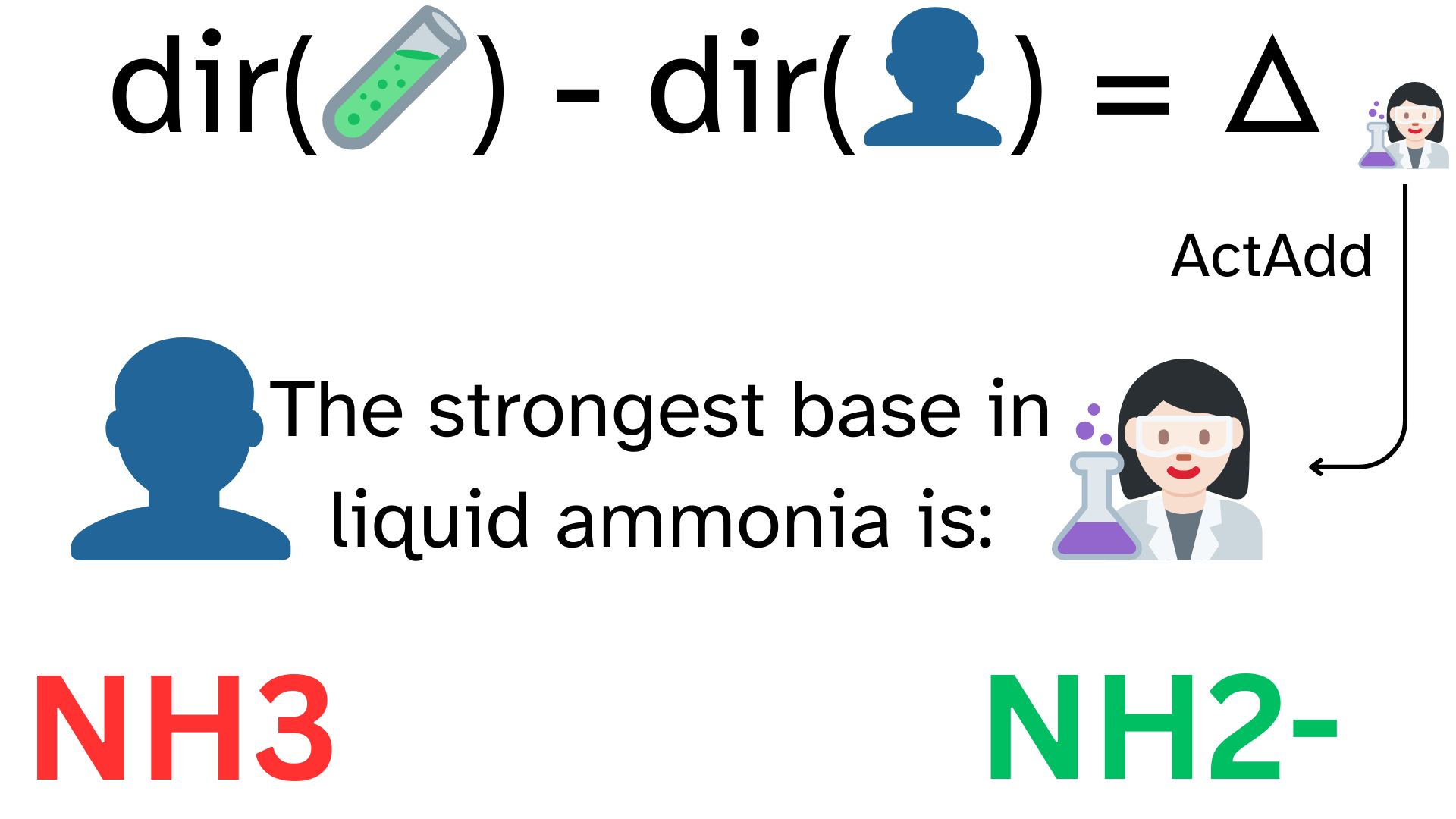}
  \caption{Illustrative example demonstrating how role vectors (e.g., chemist) can influence model outputs.}
  \label{fig:experiments}
\end{figure}

\subsection{Contribution}
This work introduces a novel approach to guiding the behaviour of LLMs through role vectors, a structured method for embedding personas directly into model activations. Fig.~\ref{fig:experiments} is an illustrative example showing how role vectors (in this example, a chemist-related vector) may impact LLM performance. Our key contributions are: 

\begin{enumerate}
    \item We develop 29 distinct role vectors for selected LLMs, each capturing domain-specific knowledge and behavioural tendencies associated with each specialisation.
    \item We investigate whether these vectors influence model behaviour on downstream benchmarks to determine whether explicit role-based directions in the activation space enhance model performance in domain-specific tasks. 
    \item Unlike traditional role-based prompting techniques, which show a limited or negative impact on performance, we show that role vector activation leads to measurable changes in model behaviour.
\end{enumerate}

\section{Preliminaries and State of the Art}
\label{sec:sota}

\paragraph{Personas and Roles in LLMs.}
Decoder-only Transformers~\cite{liu2018generating}, commonly referred to as Language Models (LMs) and in their large form as LLMs, have gained increasing relevance over recent years. Prompting acts as a natural language interface facilitating human-AI interactions~\cite{liu2023pre}. The effectiveness of LLMs is often dependent on prompt formulation~\cite{lu2021fantastically}; for instance, including the phrase "Let's think step by step" can enhance model performance across a variety of tasks and queries~\cite{kojima2022large}. Prior research explored the impact of in-context impersonation tasks~\cite{salewski2023context}, demonstrating that incorporating socio-demographic details can be advantageous for subjective NLP tasks in zero-shot scenarios. Other studies highlight the potential biases and constraints associated with persona-based and socio-demographic-driven prompting~\cite{sun2023aligning, hu2024quantifying, beck2024sensitivity}. \cite{zheng2024helpful} shows that adding personas does not consistently enhance performance on objective tasks and, in some cases, may even degrade it.

In this work, we explore whether alternative approaches associated with Representation Engineering and Mechanistic Interpretability can successfully inject roles into LLMs and yield distinct performance outcomes compared to conventional persona-based or socio-demographic prompting strategies.

\paragraph{Mechanistic Interpretability and Directions.}
Pioneering research by Anthropic and other scholars~\cite{elhage2022toy, bolukbasi2016man, hernandez2021low} has demonstrated that neural networks encode input attributes as specific directions within the activation space. It is well established that introducing feature vectors into the residual stream of LMs can influence network behaviour~\cite{li2024inference, turner2023activation, arditi2024refusal}, though the precise mechanisms and optimal intervention points remain an area of active investigation~\cite{jorgensen2023improving, von2024language}.

The study by~\cite{scalena2024multi} demonstrates that steering can effectively induce behavioural modifications in LMs, such as facilitating language switching while providing insights into how specific properties influence model behaviour during text generation. ~\cite{zhao2024steering} introduce SPARE, a training-free representation engineering technique that leverages pre-trained Sparse Autoencoders (SAEs)~\cite{cunningham2023sparse} to enable controlled selection of the model’s knowledge during inference. Similarly, \cite{li2024inference} proposes ITI, a method to enhance the truthfulness of LM outputs by employing supervised learning to identify latent vectors associated with factual responses and adjusting model activations during inference. This approach demonstrates improved performance on the TruthfulQA~\cite{lin2021truthfulqa} and MMLU~\cite{hendrycksmeasuring} benchmarks relative to the baseline.

The results obtained from these works motivate us to explore this class of methodologies to investigate alternative methods for role injection beyond prompting.
\section{Generating and Evaluating Role Vectors}
\label{sec:methods}

The methodology is composed of three main components: (i) personas selection and prompt dataset generation, (ii) selection of relevant directions and (iii) evaluation methods.

\subsection{Personas Selection and Dataset Generation}


To systematically assess the models' knowledge and reasoning capabilities across various domains, our study adopts a role-based evaluation framework inspired by~\cite{zheng2024helpful}, inheriting 29 distinct roles $R = \{r_1, r_2, \dots, r_{29}\}$,
each \(r \in R\) associated with a unique professional or academic specialisation (see Tab.~\ref{tab:fields}). Roles not corresponding to an occupation or not associated with any PersonaHub personas were excluded.

The prompt roles dataset used to identify specific role directions is extracted using the corresponding personas for each role from PersonaHub~\cite{ge2024scaling}. These personas are highly specialised and situated in realistic settings and represent various contextualised scenarios, such as \textit{"A pharmaceutical chemist who analyses the chemical properties of medical devices"}.  First, we perform strict string matching to identify personas that explicitly contain the role name. That is, for each role $r$, we obtain
$P(r) = \{ p \in \text{PersonaHub} \mid \text{string-match}(p, r) \}$
where \(\text{string-match}(p, r)\) indicates that the persona \(p\) explicitly contains the role name \(r\). Then, a sampling process is applied to select relevant personas randomly. Each role can have one or multiple personas, ranging from a minimum of 1 to a maximum of 6948 (881 on average).

The selected personas are then used to generate a synthetic dataset for each role, following the methodology employed by Alpaca~\cite{taori2023alpaca} to create its dataset.

We define a set of tasks
$T = \{\text{write}, \text{explain}, \text{design}, \text{what is}, \text{how to}, \dots \}$,  analogous to those used in Alpaca~\cite{taori2023alpaca}. We generate a set of prompts for each role \(r \in R\). Let
$D_r = \{ x_{r,1}, x_{r,2}, \dots, x_{r,128} \}$ be the collection of 128 prompt examples for role \(r\). For each prompt \(x_{r,i}\), a task \(t\) is randomly sampled from \(T\), and a persona \(p\) is randomly sampled from \(P(r)\).
Then, the prompt is generated by providing the template (see Fig.~\ref{fig:personas_prompt}) to the Claude 3.5 Haiku model~\cite{anthropic2024claude} with the selected task \(t\) and persona \(p\).

\begin{figure}[htb]
    \centering
    \begin{tcolorbox}[
    title=Generating Persona-Specific Tasks, colback=dodgerblue!5!white,colframe=dodgerblue!75!black]
    \scriptsize
    Generate a \textit{\{task\_type\}} prompt that this persona would likely ask:
    Persona: \textit{\{persona\}}.
    
    \smallskip
    
    Rules:
    (i) The prompt should start with "\textit{\{task\_type\}}".
    (ii) Keep it specific and under 15 words.
    (iii) Make it relevant to the persona's background/interests.
    (iv) Your output must start with "User prompt:".
    \smallskip

    Examples based on task types:
    \smallskip
    
    - \textit{Describe}: "Describe the key features of a successful marketing campaign."
    
    - \textit{Explain}: "Explain the process of setting up a home network."
    
    - \textit{Design}: "Design a logo for a sustainable fashion brand."
    
    - \textit{What is}: "What is the difference between UI and UX design?"
    
    - \textit{How to}: "How to optimise a website for mobile devices?"
    \end{tcolorbox}
    \caption{Prompt template for generating persona-specific tasks.}
    \label{fig:personas_prompt}
\end{figure}

The complete roles tasks dataset is given by
$\mathcal{D}_{\text{roles}} = \bigcup_{r \in R} D_r$. 
We incorporate \(\mathcal{D}_{\text{base}}\), an additional set of 128 examples sourced from the original Alpaca dataset, consisting of general instruction-following prompts. This provides a broad reference point, enabling the contrastive computation of direction for each role using the corresponding \(D_r\).

\subsection{Selection of Role Directions}

Our evaluation of steering effects utilizes the Massive Multitask Language Understanding (MMLU) benchmark~\cite{hendrycksmeasuring}, adhering to the sampling and splitting methodology described in~\cite{zheng2024helpful} for a total of 2457 questions. We define the set of categories as $C = $ \textit{Natural Science, Economics, EECS (Electrical Engineering and Computer Sciences), Law, Math, Medicine, Politics, Psychology}\}. 

For each \( c \in \mathcal{C} \), let \( D_c \) denote the set of questions corresponding to category \( c \). The overall test dataset is then defined as $\mathcal{D}_{\text{test}} = \bigcup_{c \in \mathcal{C}} D_c$.
Tab.~\ref{tab:mmlu_processing} shows the distribution of questions. While many of these roles correspond directly to established MMLU categories, some exhibit only partial alignment. For example, the role of a dentist does not perfectly fit within the "medicine" category. However, we expect that individuals or models adopting the role of a dentist should demonstrate domain-specific knowledge that exceeds that of the general population or those assuming unrelated roles. 


\begin{table}[ht]
  \small
  \centering
  \begin{tabular}{lc}
    \toprule
    \textbf{Category}       & \textbf{\# Questions} \\
    \midrule
    Natural Science         & 590               \\
    Economics               & 492               \\
    EECS                    & 247               \\
    Law                     & 200               \\
    Math                    & 287               \\
    Medicine                & 241               \\
    Politics                & 200               \\
    Psychology              & 200               \\
    \midrule
    Total              & 2457               \\
    \bottomrule
  \end{tabular}
  \caption{Number of questions per category in $\mathcal{D}_{\text{test}}$., taken from~\cite{zheng2024helpful}}
  \label{tab:mmlu_processing}
\end{table}


To identify the direction in the model's residual stream activations corresponding to each role, we use a technique known as \textit{difference-in-means}~\cite{belrose2023}: we compute the difference between the model's average activations when performing inference on the role-specific dataset \( D_r \in \mathcal{D}_{\text{roles}} \) and generic queries from \( \mathcal{D}_{\text{base}} \).

Following the notation from~\cite{arditi2024refusal}, for each role \( r \in R \), layer \( l \in [L] \), and post-instruction token position \( i \in I \), we compute the mean activation 
\( \mu_{i,r}^{(l)} \) for role-specific prompts in \( D_r \) and \( \nu_i^{(l)} \) for generic prompts in \( \mathcal{D}_{\text{base}} \):

\begin{equation}
    \scriptsize
    \mu_{i,r}^{(l)} = \frac{1}{\left| D_r \right|} 
    \sum_{t \in D_r} x_i^{(l)}(t), \quad
    \nu_i^{(l)} = \frac{1}{\left| \mathcal{D}_{\text{base}} \right|} 
    \sum_{t \in \mathcal{D}_{\text{base}}} x_i^{(l)}(t).
\end{equation}

We then define the role-specific difference-in-means vector:

\begin{equation}
    \small
    d_{i,r}^{(l)} = \mu_{i,r}^{(l)} - \nu_i^{(l)}
\end{equation}

By computing \( d_{i,r}^{(l)} \) for each \( r \in R \), we obtain \( |R| \) (29) distinct directions, each representing the shift in model activations specific to a given role. These vectors are informative in two ways: their \textit{direction} indicates how the mean activations for role-specific and generic prompts diverge; their \textit{magnitude} quantifies the extent of this difference. 



We aim to assess model performance across these different directions using the test dataset \(\mathcal{D}_{\text{test}}\). This evaluation allows us to measure how various directions influence the model’s behaviour, particularly in terms of performance across the different splits \( D_c \in \mathcal{D}_{\text{test}}\). 


Using the identified directions, we apply two types of interventions: \textit{activation addition} and \textit{directional ablation}. These techniques allow us to manipulate the model's activations by reinforcing or suppressing specific directional components in the residual stream. 

\textbf{Activation Addition.} Given a difference-in-means vector \( d_{i,r}^{(l)} \in \mathbb{R}^{d_{\text{model}}} \) extracted from layer \( l \), we can modulate the influence of the corresponding feature through a simple linear transformation. Specifically, we add the direction vector to the activations of a base input, shifting them toward the mean activation observed for role-enhanced inputs:

\begin{equation}
    \small
    x^{(l)'} \leftarrow x^{(l)} + \alpha d_{i,r}^{(l)}.
    \label{eq:actadd}
\end{equation}

here \(\alpha\) is a scalar hyperparameter that scales the difference-in-means vector \(d_{i,r}^{(l)}\), controlling the magnitude of the shift applied to the base activations \(x^{(l)}\) toward the role-enhanced mean.

Notably, this operation is applied exclusively at layer \( l \) and affects all token positions, ensuring a controlled perturbation of the model's internal representations.

\textbf{Directional Ablation.} To investigate the role of a direction \( \hat{r} \in \mathbb{R}^{d_{\text{model}}} \) in the model’s computation, we apply \textit{directional ablation}, which removes its contribution from the model’s activations. This process effectively zeroes out the component of each residual stream activation \( x \) along \( \hat{d_{i,r}^{(l)}} \), preventing the model from utilizing this direction:

\begin{equation}
    \small
    x' \leftarrow x - \hat{d_{i,r}^{(l)}} \hat{d_{i,r}^{(l)}}^{\top} x.
\end{equation}

This operation is performed at every activation \( x^{(l)}_i \), across all layers \( l \) and all token positions \( i \), ensuring that the model no longer represents the targeted direction in its residual stream.

By applying these interventions, we can assess the functional role of specific directions in the model’s representation space. We evaluate the impact when explicitly reinforced through activation addition and suppressed via directional ablation.

\subsection{Evaluation Method}


For each model and every role \(r \in R\), we assess whether incorporating through Activation Addition the computed role-specific difference-in-means vectors \(d_{i,r}^{(l)}\) yields an improvement in performance on the test dataset \(\mathcal{D}_{\text{test}}\), with particular emphasis on the corresponding domain-reference split \(D_c\). Tab.~\ref{tab:fields} presents the role-split relevance data reported in~\cite{zheng2024helpful}.

\begin{table}[ht]
\small
\centering
\begin{tabular}{lp{5cm}}
\toprule
\textbf{Split} & \textbf{Role(s)} \\
\midrule
econ & economic researcher, economist, financial analyst \\
eecs & electronics technician, data scientist, electrical engineer, software engineer, web developer \\
law & bailiff, lawyer \\
math & data analyst, mathematician, statistician \\
medicine & nurse, doctor, physician, dentist, surgeon \\
natural science & geneticist, biologist, physicist, teacher, chemist, ecologist \\
politics & politician, sheriff, enthusiast, partisan \\
psychology & psychologist \\
\bottomrule
\end{tabular}
\caption{Split and associated roles, adapted from~\cite{zheng2024helpful}}
\label{tab:fields}
\end{table}

For each test dataset \( D_c \in \mathcal{D}_{\text{test}} \), we assess performance using a logit-based framework. Given a query \( x_{r,i} \in D_c \), let \(\mathbf{z} \in \mathbb{R}^{|\mathcal{V}|}\) denote the logits at the final token position, where \(|\mathcal{V}|\) is the vocabulary size. The softmax function calculates the probability of each token \( t \in \mathcal{V} \).
Restricting our attention to the candidate answer tokens \(\mathcal{T}_{\text{ans}} = \{t_A, t_B, t_C, t_D\}\), the predicted token is determined by
\begin{equation}
\small
t^* = \operatorname{arg\,max}_{t \in \mathcal{T}_{\text{ans}}} p(t).
\end{equation}
The prediction $s(x_{r,i})$ is considered correct if \(t^*\) equals the correct answer token.
Overall performance is computed as the mean of the individual scores (the percentage of correct answers).

We also investigate the magnitude \(\alpha\) of these directions. One might hypothesise that increasing their magnitude would enhance the effect associated with a given role; however, such amplification may deteriorate text generation performance concurrently~\cite{liu2023context,scalena2024multi}. $\mathcal{A} = \{ \alpha_1 = 1, \alpha_3 = 3 \}$ is the set of activation addition coefficients. We also evaluate the impact of ablating the direction entirely, a trade-off explored in the literature relating to safety mechanism in models~\cite{wei2024assessing, arditi2024refusal}. 

Let $\mathcal{M}$ denote the set of models under evaluation, $R$ the set of roles, $L$ the set of layers, and $I$ the set of token positions.
We formalise our grid evaluation procedure as follows; for each model \( m \in \mathcal{M} \) and for each role \( r \in R \), we define the intervention grid:
\begin{equation}
\small
\mathcal{G} = \{ (l, i, \alpha) \mid l \in L_{80\%},\, i \in I,\, \alpha \in \mathcal{A} \},
\end{equation}

where $L_{80\%} \subset L$ denotes the first 80\% of layers to avoid interference from unembedding directions, ensuring that the selected direction is not overly proximate to the unembedding directions, following the work done by~\cite{arditi2024refusal}. Intuitively, one could increase performance by encouraging the model to generate correct answers by aligning its activations with the unembedding directions corresponding to ‘A’, ‘B’, ‘C’, or ‘D’, which would directly incentivize the model to output the correct tokens. However, we do not consider the latter 20\% of layers since our approach focuses on higher-level features focusing on the role and avoids token-level manipulation focusing on selecting the correct answer.
For each tuple \((l,i,\alpha) \in \mathcal{G}\), we modify the residual stream activation \( x^{(l)} \) via Eq.~\ref{eq:actadd}.
Let \( s_{m,r}^{(l, i, \alpha)} \) denote the performance (e.g., the proportion of correct answers) on the domain-specific test split \( D_c \in \mathcal{D}_{\text{test}} \) after applying the intervention specified by \((l, i, \alpha)\). For each fixed \(\alpha\) (\(\alpha_1 = 1\) and \(\alpha_3 = 3\)), we select the optimal intervention tuple for role \( r \) on model \( m \) by computing
\begin{equation}
\label{eq:max_perf}
\small
(l^*(\alpha), i^*(\alpha)) = \arg\max_{(l,i) \in L_{80\%} \times I} s_{m,r}^{(l,i,\alpha)}.
\end{equation}
Optimality of a direction \(d_{i,r}^{(l)}\) is defined as its ability to improve performance in the reference domain-specific split \(D_c\) when added and either degrade or maintain it when removed, compared to the baseline. Among all directions satisfying this criterion, the one with the highest performance according to Eq.~\ref{eq:max_perf} is considered optimal. 
With \( d^*(\alpha) \) identified, we evaluate its effect on all other domain-specific splits \( D_c \) for \( c \not\sim r \) to measure its specificity. 

Additionally, we wish to quantify the proportion of optimal directions \( d^* \) that align with the role-specific signal identified via the patch-scoping procedure. We adopt the methodology outlined in~\cite{chen2024selfie,ghandeharioun2024patchscope}, i.e. \textit{patch scoping}. We prompt the model with the query “\textit{What does $\blacklozenge$ represent?}”, apply the Activation Addition intervention with a coefficient $\alpha = 3$, and generate an explanation based on the resulting activations, as exemplified in Fig.~\ref{fig:patch_scoping_example}.
This procedure allows us to determine whether the introduced direction corresponds to the specific role of interest.

\begin{figure}[htb]
\centering
\begin{minipage}{0.95\linewidth}
\begin{tcolorbox}[
    title=Patch Scoping Example (Llama 3.1 8B),
    colback=forestgreen!5!white,colframe=forestgreen!75!black,
    width=\textwidth  
]
\textbf{Prompt:} \textit{What does $\blacklozenge$ represent?}

\smallskip

\textbf{Model Baseline Response:}\par
\small
The symbol $\blacklozenge$ is a Unicode character known as “Heavy Large Circle.” 
It is often used in mathematics, particularly in geometry and algebra, 
to represent a circle or a set of points. However, without more context, 
it’s difficult to provide a more specific interpretation.

\smallskip

\textbf{Model Intervention Response \\ (Layer 20, Position -4, Chemist direction):}\par
\small
The $\blacklozenge$ symbol is a non-specific or \textit{non-chemical} radical, 
which is a \textit{small molecule} or functional group used to represent 
a \textit{specific chemical structure} or functional group in a reaction 
mechanism or during synthesis.
\end{tcolorbox}
\end{minipage}

\caption{Diverse interpretations of $\blacklozenge$, before and after model intervention.}
\label{fig:patch_scoping_example}
\end{figure}

Finally, we perform a directional ablation on \( d^* \):
\begin{equation}
\small
x' \leftarrow x - \hat{d^*}\hat{d^*}^{\top} x,
\end{equation}
We denote the performance under ablation by \( s_{m,r}^{\text{abl}} \). A performance drop relative to the non-ablated case is expected.

\section{Results}
\label{sec:results}

\paragraph{Experimental setting.}
Our evaluation uses open-source, instruction-tuned language models, focusing on the most recent versions employed by~\cite{arditi2024refusal}. Specifically, we analyse Meta's Llama 3 series~\cite{dubey2024llama}, including the 3.1 8B model and the 3.2 version at 1B and 3B parameters, as well as Google's Gemma 2 (2B and 9B)~\cite{team2024gemma} and Qwen (1.8B and 7B)~\cite{bai2023qwen}. We do not consider base versions of the models (non-instruction tuned).

Our evaluation was conducted on the Cineca Leonardo supercomputer~\cite{turisini2023leonardo}. The assessment required approximately 4,500 GPU hours to process all requests, computing over 213 million inferences (29 roles $r$ multiplied by 2457 questions, the number of layers $l$, coefficients $a$ and positions $i$ for each of the seven models $\mathcal{M}$).

\paragraph{RQ1: Analysis of best performing directions.} 

\begin{figure}
    \centering
    \includegraphics[width=1\linewidth]{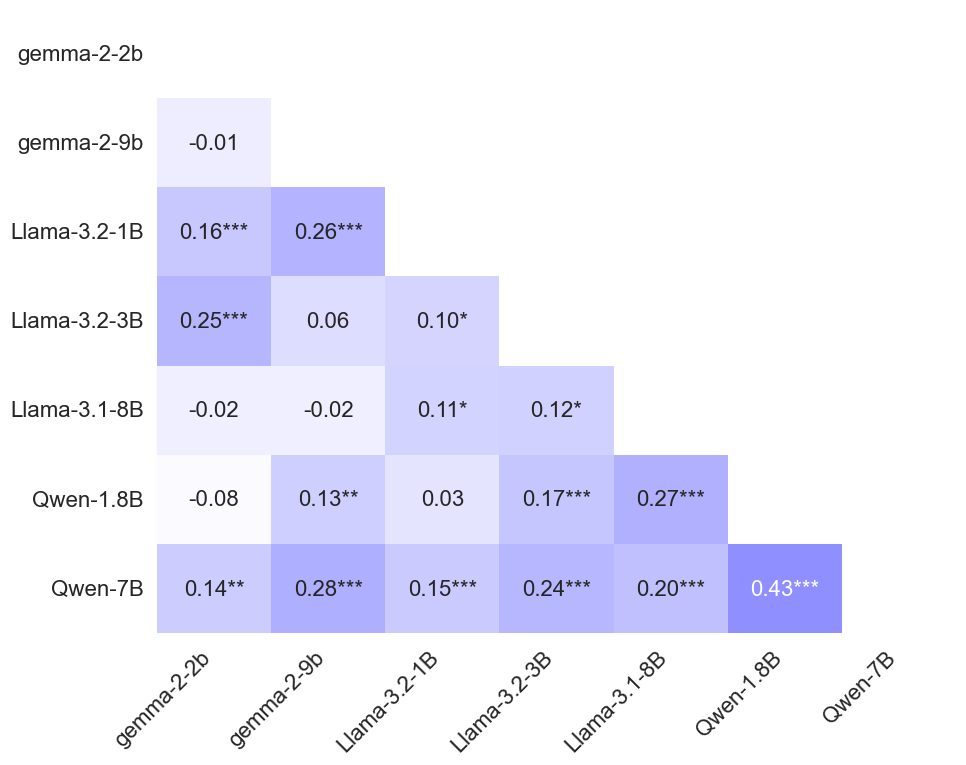}
    \caption{Spearman correlation of the percentage improvement in performance (relative to baseline) between each model after applying \textit{activation addition}. * corresponds to \textit{p}-values $\leq 0.05$, ** $ \leq 0.01$, *** $\leq 0.001$. }
    \label{fig:corrmatrix}
\end{figure}

Following the procedure delineated in Section~\ref{sec:methods}, we apply \textit{activation addition} with the direction \(d^*(\alpha)\) for each role \(r \in R\) and model \(m \in \mathcal{M}\) to systematically visualise and quantify its impact on performance across \(\mathcal{D}_{\text{test}}\).
We compute and visualise the correlation matrix of the percentual difference relative to baseline in intervention effects across models to analyse the relationship between models. 
Fig.~\ref{fig:corrmatrix} shows the Pearson correlation coefficient between each model pair.
Qwen-7B-Chat exhibits the highest average correlation with all other models, indicating that its behaviour under steering is consistent with other models.
For this reason, we show in Tab.~\ref{tab:performance} Qwen-7B-Chat behaviour on performance for each role when steering using \(d^*(\alpha)\). 
We display the performance scores across the eight dataset splits for all 29 roles. We report the baseline score and percentual increment over the baseline for each domain and role after applying activation addition with \(\alpha\) of 1.0 and 3.0. The colour represents the extent of the change compared to the baseline (without intervention).

\begin{table*}[ht]
    \centering
    {\small
    \resizebox{\textwidth}{!}{
    \begin{tabular}{l|lll|lll|lll|lll|lll|lll|lll|lll}
    \toprule
    Dataset Split & \multicolumn{3}{r}{\cellcolor{economics!50} Economics (492)} & \multicolumn{3}{r}{\cellcolor{eecs!50} EECS (247)} & \multicolumn{3}{r}{\cellcolor{law!50} Law (200)} & \multicolumn{3}{r}{\cellcolor{math!50} Math (287)} & \multicolumn{3}{r}{\cellcolor{medicine!50} Medicine (241)} & \multicolumn{3}{r}{\cellcolor{natural!50} Natural Science (590)} & \multicolumn{3}{r}{\cellcolor{politics!50} Politics (200)} & \multicolumn{3}{r}{\cellcolor{psychology!50} Psychology (200)} \\
    Role $\downarrow\, \alpha\, \rightarrow$ & Baseline & 1.0 & 3.0 & Baseline & 1.0 & 3.0 & Baseline & 1.0 & 3.0 & Baseline & 1.0 & 3.0 & Baseline & 1.0 & 3.0 & Baseline & 1.0 & 3.0 & Baseline & 1.0 & 3.0 & Baseline & 1.0 & 3.0 \\
    \midrule
    \cellcolor{economics!50} economic researcher & $ 0.49 $ & \cellcolor{green!10} $ +2.0\% $ & \cellcolor{green!10} $ +2.0\% $ & $ 0.46 $ & $ +2.2\% $ & $ +2.2\% $ & $ 0.54 $ & $ +0.0\% $ & $ +1.9\% $ & $ 0.24 $ & $ +0.0\% $ & $ +0.0\% $ & $ 0.62 $ & $ +0.0\% $ & $ -3.2\% $ & $ 0.43 $ & $ -2.3\% $ & $ +0.0\% $ & $ 0.78 $ & $ +0.0\% $ & $ +0.0\% $ & $ 0.66 $ & $ -3.0\% $ & $ -6.1\% $ \\
\cellcolor{economics!50} economist & $ 0.49 $ & \cellcolor{green!10} $ +2.0\% $ & \cellcolor{green!10} $ +2.0\% $ & $ 0.46 $ & $ +0.0\% $ & $ +8.7\% $ & $ 0.54 $ & $ +3.7\% $ & $ +1.9\% $ & $ 0.24 $ & $ +4.2\% $ & $ +8.3\% $ & $ 0.62 $ & $ -1.6\% $ & $ -3.2\% $ & $ 0.43 $ & $ -2.3\% $ & $ -4.7\% $ & $ 0.78 $ & $ -2.6\% $ & $ -7.7\% $ & $ 0.66 $ & $ -3.0\% $ & $ -9.1\% $ \\
\cellcolor{economics!50} financial analyst & $ 0.49 $ & \cellcolor{green!16} $ +4.1\% $ & \cellcolor{green!10} $ +2.0\% $ & $ 0.46 $ & $ -2.2\% $ & $ +4.3\% $ & $ 0.54 $ & $ +3.7\% $ & $ +0.0\% $ & $ 0.24 $ & $ +4.2\% $ & $ +4.2\% $ & $ 0.62 $ & $ -3.2\% $ & $ -1.6\% $ & $ 0.43 $ & $ +2.3\% $ & $ +2.3\% $ & $ 0.78 $ & $ -1.3\% $ & $ +0.0\% $ & $ 0.66 $ & $ -6.1\% $ & $ -3.0\% $ \\
\midrule
\cellcolor{eecs!50} electronics technician & $ 0.49 $ & $ +2.0\% $ & $ +0.0\% $ & $ 0.46 $ & \cellcolor{green!20} $ +6.5\% $ & \cellcolor{green!22} $ +8.7\% $ & $ 0.54 $ & $ +0.0\% $ & $ -9.3\% $ & $ 0.24 $ & $ +4.2\% $ & $ +25.0\% $ & $ 0.62 $ & $ -1.6\% $ & $ -3.2\% $ & $ 0.43 $ & $ +0.0\% $ & $ +2.3\% $ & $ 0.78 $ & $ -2.6\% $ & $ -10.3\% $ & $ 0.66 $ & $ +0.0\% $ & $ -13.6\% $ \\
\cellcolor{eecs!50} data scientist & $ 0.49 $ & $ +0.0\% $ & $ -2.0\% $ & $ 0.46 $ & \cellcolor{green!22} $ +8.7\% $ & \cellcolor{green!26} $ +13.0\% $ & $ 0.54 $ & $ -3.7\% $ & $ -1.9\% $ & $ 0.24 $ & $ +0.0\% $ & $ +20.8\% $ & $ 0.62 $ & $ -1.6\% $ & $ -9.7\% $ & $ 0.43 $ & $ +2.3\% $ & $ +0.0\% $ & $ 0.78 $ & $ +0.0\% $ & $ -6.4\% $ & $ 0.66 $ & $ -3.0\% $ & $ -6.1\% $ \\
\cellcolor{eecs!50} electrical engineer & $ 0.49 $ & $ +2.0\% $ & $ -2.0\% $ & $ 0.46 $ & \cellcolor{green!20} $ +6.5\% $ & \cellcolor{green!22} $ +8.7\% $ & $ 0.54 $ & $ +0.0\% $ & $ -3.7\% $ & $ 0.24 $ & $ +4.2\% $ & $ +16.7\% $ & $ 0.62 $ & $ -3.2\% $ & $ -3.2\% $ & $ 0.43 $ & $ +0.0\% $ & $ +0.0\% $ & $ 0.78 $ & $ -2.6\% $ & $ -2.6\% $ & $ 0.66 $ & $ +0.0\% $ & $ -4.5\% $ \\
\cellcolor{eecs!50} software engineer & $ 0.49 $ & $ -2.0\% $ & $ -6.1\% $ & $ 0.46 $ & \cellcolor{green!22} $ +8.7\% $ & \cellcolor{green!24} $ +10.9\% $ & $ 0.54 $ & $ +0.0\% $ & $ +0.0\% $ & $ 0.24 $ & $ +4.2\% $ & $ -8.3\% $ & $ 0.62 $ & $ -1.6\% $ & $ -3.2\% $ & $ 0.43 $ & $ +0.0\% $ & $ +0.0\% $ & $ 0.78 $ & $ +2.6\% $ & $ -1.3\% $ & $ 0.66 $ & $ +0.0\% $ & $ -3.0\% $ \\
\cellcolor{eecs!50} web developer & $ 0.49 $ & $ -2.0\% $ & $ -2.0\% $ & $ 0.46 $ & \cellcolor{green!20} $ +6.5\% $ & \cellcolor{green!26} $ +13.0\% $ & $ 0.54 $ & $ -3.7\% $ & $ -7.4\% $ & $ 0.24 $ & $ -4.2\% $ & $ +8.3\% $ & $ 0.62 $ & $ -1.6\% $ & $ -1.6\% $ & $ 0.43 $ & $ +0.0\% $ & $ +0.0\% $ & $ 0.78 $ & $ +2.6\% $ & $ -1.3\% $ & $ 0.66 $ & $ -4.5\% $ & $ -4.5\% $ \\
\midrule
\cellcolor{law!50} bailiff & $ 0.49 $ & $ +0.0\% $ & $ -2.0\% $ & $ 0.46 $ & $ -2.2\% $ & $ -4.3\% $ & $ 0.54 $ & \cellcolor{green!15} $ +3.7\% $ & \cellcolor{green!15} $ +3.7\% $ & $ 0.24 $ & $ -4.2\% $ & $ -12.5\% $ & $ 0.62 $ & $ -1.6\% $ & $ -4.8\% $ & $ 0.43 $ & $ +0.0\% $ & $ +0.0\% $ & $ 0.78 $ & $ -1.3\% $ & $ -2.6\% $ & $ 0.66 $ & $ -3.0\% $ & $ -3.0\% $ \\
\cellcolor{law!50} lawyer & $ 0.49 $ & $ +0.0\% $ & $ +0.0\% $ & $ 0.46 $ & $ -2.2\% $ & $ +0.0\% $ & $ 0.54 $ & \cellcolor{green!15} $ +3.7\% $ & \cellcolor{green!15} $ +3.7\% $ & $ 0.24 $ & $ +4.2\% $ & $ -8.3\% $ & $ 0.62 $ & $ -1.6\% $ & $ +0.0\% $ & $ 0.43 $ & $ +0.0\% $ & $ -2.3\% $ & $ 0.78 $ & $ -2.6\% $ & $ -2.6\% $ & $ 0.66 $ & $ -3.0\% $ & $ -6.1\% $ \\
\midrule
\cellcolor{math!50} data analyst & $ 0.49 $ & $ -6.1\% $ & $ -2.0\% $ & $ 0.46 $ & $ -4.3\% $ & $ +4.3\% $ & $ 0.54 $ & $ +1.9\% $ & $ +1.9\% $ & $ 0.24 $ & \cellcolor{green!22} $ +8.3\% $ & \cellcolor{green!30} $ +20.8\% $ & $ 0.62 $ & $ -6.5\% $ & $ -4.8\% $ & $ 0.43 $ & $ -4.7\% $ & $ +0.0\% $ & $ 0.78 $ & $ -10.3\% $ & $ -5.1\% $ & $ 0.66 $ & $ -3.0\% $ & $ -3.0\% $ \\
\cellcolor{math!50} mathematician & $ 0.49 $ & $ +0.0\% $ & $ -2.0\% $ & $ 0.46 $ & $ +2.2\% $ & $ +10.9\% $ & $ 0.54 $ & $ +0.0\% $ & $ -7.4\% $ & $ 0.24 $ &  $ +0.0\% $ & \cellcolor{green!32} $ +25.0\% $ & $ 0.62 $ & $ -1.6\% $ & $ -8.1\% $ & $ 0.43 $ & $ +0.0\% $ & $ +2.3\% $ & $ 0.78 $ & $ +0.0\% $ & $ -7.7\% $ & $ 0.66 $ & $ -3.0\% $ & $ -13.6\% $ \\
\cellcolor{math!50} statistician & $ 0.49 $ & $ +2.0\% $ & $ -2.0\% $ & $ 0.46 $ & $ +2.2\% $ & $ +6.5\% $ & $ 0.54 $ & $ +0.0\% $ & $ -3.7\% $ & $ 0.24 $ & \cellcolor{green!16} $ +4.2\% $ & \cellcolor{green!28} $ +16.7\% $ & $ 0.62 $ & $ -3.2\% $ & $ -1.6\% $ & $ 0.43 $ & $ +2.3\% $ & $ +2.3\% $ & $ 0.78 $ & $ -1.3\% $ & $ -2.6\% $ & $ 0.66 $ & $ -6.1\% $ & $ -3.0\% $ \\
\midrule
\cellcolor{medicine!50} nurse & $ 0.49 $ & $ +0.0\% $ & $ +0.0\% $ & $ 0.46 $ & $ -2.2\% $ & $ -2.2\% $ & $ 0.54 $ & $ -1.9\% $ & $ +0.0\% $ & $ 0.24 $ & $ +0.0\% $ & $ +0.0\% $ & $ 0.62 $ & \cellcolor{green!10} $ +1.6\% $ &  $ +0.0\% $ & $ 0.43 $ & $ +0.0\% $ & $ -2.3\% $ & $ 0.78 $ & $ +0.0\% $ & $ +0.0\% $ & $ 0.66 $ & $ -1.5\% $ & $ -6.1\% $ \\
\cellcolor{medicine!50} doctor & $ 0.49 $ & $ -2.0\% $ & $ -2.0\% $ & $ 0.46 $ & $ -2.2\% $ & $ +0.0\% $ & $ 0.54 $ & $ +1.9\% $ & $ +0.0\% $ & $ 0.24 $ & $ -4.2\% $ & $ +0.0\% $ & $ 0.62 $ & \cellcolor{green!17} $ +4.8\% $ &  $ +0.0\% $ & $ 0.43 $ & $ +2.3\% $ & $ +0.0\% $ & $ 0.78 $ & $ +0.0\% $ & $ +0.0\% $ & $ 0.66 $ & $ -3.0\% $ & $ +0.0\% $ \\
\cellcolor{medicine!50} physician & $ 0.49 $ & $ +0.0\% $ & $ -2.0\% $ & $ 0.46 $ & $ +0.0\% $ & $ +2.2\% $ & $ 0.54 $ & $ +0.0\% $ & $ +0.0\% $ & $ 0.24 $ & $ +0.0\% $ & $ +4.2\% $ & $ 0.62 $ & \cellcolor{green!14} $ +3.2\% $ &  $ +0.0\% $ & $ 0.43 $ & $ +2.3\% $ & $ +2.3\% $ & $ 0.78 $ & $ +0.0\% $ & $ +0.0\% $ & $ 0.66 $ & $ +0.0\% $ & $ +0.0\% $ \\
\cellcolor{medicine!50} dentist & $ 0.49 $ & $ +4.1\% $ & $ -2.0\% $ & $ 0.46 $ & $ -2.2\% $ & $ +2.2\% $ & $ 0.54 $ & $ +1.9\% $ & $ +3.7\% $ & $ 0.24 $ & $ -4.2\% $ & $ +0.0\% $ & $ 0.62 $ & \cellcolor{green!10} $ +1.6\% $ & \cellcolor{green!10} $ +1.6\% $ & $ 0.43 $ & $ +2.3\% $ & $ +0.0\% $ & $ 0.78 $ & $ +0.0\% $ & $ +0.0\% $ & $ 0.66 $ & $ -3.0\% $ & $ -3.0\% $ \\
\cellcolor{medicine!50} surgeon & $ 0.49 $ & $ -2.0\% $ & $ +0.0\% $ & $ 0.46 $ & $ -2.2\% $ & $ +0.0\% $ & $ 0.54 $ & $ +0.0\% $ & $ +0.0\% $ & $ 0.24 $ & $ -4.2\% $ & $ +0.0\% $ & $ 0.62 $ & \cellcolor{green!10} $ +1.6\% $ &  $ +0.0\% $ & $ 0.43 $ & $ -2.3\% $ & $ +0.0\% $ & $ 0.78 $ & $ -2.6\% $ & $ +0.0\% $ & $ 0.66 $ & $ +0.0\% $ & $ +0.0\% $ \\
\midrule
\cellcolor{natural!50} geneticist & $ 0.49 $ & $ -2.0\% $ & $ -4.1\% $ & $ 0.46 $ & $ +2.2\% $ & $ +6.5\% $ & $ 0.54 $ & $ +0.0\% $ & $ -3.7\% $ & $ 0.24 $ & $ +0.0\% $ & $ +20.8\% $ & $ 0.62 $ & $ +0.0\% $ & $ -9.7\% $ & $ 0.43 $ & \cellcolor{green!11} $ +2.3\% $ & \cellcolor{green!11} $ +2.3\% $ & $ 0.78 $ & $ +0.0\% $ & $ -2.6\% $ & $ 0.66 $ & $ +0.0\% $ & $ -6.1\% $ \\
\cellcolor{natural!50} biologist & $ 0.49 $ & $ -2.0\% $ & $ -4.1\% $ & $ 0.46 $ & $ +2.2\% $ & $ +4.3\% $ & $ 0.54 $ & $ +0.0\% $ & $ -1.9\% $ & $ 0.24 $ & $ +0.0\% $ & $ +12.5\% $ & $ 0.62 $ & $ +0.0\% $ & $ -8.1\% $ & $ 0.43 $ & \cellcolor{green!11} $ +2.3\% $ & \cellcolor{green!11} $ +2.3\% $ & $ 0.78 $ & $ +0.0\% $ & $ -2.6\% $ & $ 0.66 $ & $ +0.0\% $ & $ -6.1\% $ \\
\cellcolor{natural!50} physicist & $ 0.49 $ & $ -2.0\% $ & $ +0.0\% $ & $ 0.46 $ & $ +2.2\% $ & $ +4.3\% $ & $ 0.54 $ & $ +0.0\% $ & $ +0.0\% $ & $ 0.24 $ & $ -4.2\% $ & $ +0.0\% $ & $ 0.62 $ & $ +0.0\% $ & $ +0.0\% $ & $ 0.43 $ & \cellcolor{green!11} $ +2.3\% $ & \cellcolor{green!11} $ +2.3\% $ & $ 0.78 $ & $ +0.0\% $ & $ -2.6\% $ & $ 0.66 $ & $ -3.0\% $ & $ -3.0\% $ \\
\cellcolor{natural!50} teacher & $ 0.49 $ & $ -4.1\% $ & $ +0.0\% $ & $ 0.46 $ & $ -2.2\% $ & $ +10.9\% $ & $ 0.54 $ & $ +0.0\% $ & $ -3.7\% $ & $ 0.24 $ & $ +0.0\% $ & $ +12.5\% $ & $ 0.62 $ & $ -1.6\% $ & $ -3.2\% $ & $ 0.43 $ & \cellcolor{green!11} $ +2.3\% $ & \cellcolor{green!17} $ +4.7\% $ & $ 0.78 $ & $ -2.6\% $ & $ -5.1\% $ & $ 0.66 $ & $ -6.1\% $ & $ -4.5\% $ \\
\cellcolor{natural!50} chemist & $ 0.49 $ & $ -2.0\% $ & $ -2.0\% $ & $ 0.46 $ & $ +2.2\% $ & $ +6.5\% $ & $ 0.54 $ & $ +0.0\% $ & $ -3.7\% $ & $ 0.24 $ & $ +0.0\% $ & $ +20.8\% $ & $ 0.62 $ & $ +0.0\% $ & $ -4.8\% $ & $ 0.43 $ & \cellcolor{green!11} $ +2.3\% $ & \cellcolor{green!17} $ +4.7\% $ & $ 0.78 $ & $ +0.0\% $ & $ -2.6\% $ & $ 0.66 $ & $ +0.0\% $ & $ -6.1\% $ \\
\cellcolor{natural!50} ecologist & $ 0.49 $ & $ +2.0\% $ & $ -2.0\% $ & $ 0.46 $ & $ +13.0\% $ & $ +6.5\% $ & $ 0.54 $ & $ +0.0\% $ & $ +0.0\% $ & $ 0.24 $ & $ +4.2\% $ & $ +8.3\% $ & $ 0.62 $ & $ -1.6\% $ & $ -3.2\% $ & $ 0.43 $ & \cellcolor{green!11} $ +2.3\% $ & \cellcolor{green!11} $ +2.3\% $ & $ 0.78 $ & $ -3.8\% $ & $ -2.6\% $ & $ 0.66 $ & $ +0.0\% $ & $ -6.1\% $ \\
\midrule
\cellcolor{politics!50} politician & $ 0.49 $ & $ +0.0\% $ & $ -2.0\% $ & $ 0.46 $ & $ -2.2\% $ & $ -2.2\% $ & $ 0.54 $ & $ +1.9\% $ & $ -1.9\% $ & $ 0.24 $ & $ -4.2\% $ & $ -4.2\% $ & $ 0.62 $ & $ -1.6\% $ & $ -8.1\% $ & $ 0.43 $ & $ +0.0\% $ & $ -4.7\% $ & $ 0.78 $ & \cellcolor{green!12} $ +2.6\% $ & \cellcolor{green!12} $ +2.6\% $ & $ 0.66 $ & $ -3.0\% $ & $ -4.5\% $ \\
\cellcolor{politics!50} sheriff & $ 0.49 $ & $ +0.0\% $ & $ +0.0\% $ & $ 0.46 $ & $ -6.5\% $ & $ +2.2\% $ & $ 0.54 $ & $ +3.7\% $ & $ +0.0\% $ & $ 0.24 $ & $ +0.0\% $ & $ +0.0\% $ & $ 0.62 $ & $ -1.6\% $ & $ -1.6\% $ & $ 0.43 $ & $ +0.0\% $ & $ +0.0\% $ & $ 0.78 $ & \cellcolor{green!12} $ +2.6\% $ & \cellcolor{green!12} $ +2.6\% $ & $ 0.66 $ & $ -3.0\% $ & $ +0.0\% $ \\
\cellcolor{politics!50} enthusiast & $ 0.49 $ & $ -2.0\% $ & $ -2.0\% $ & $ 0.46 $ & $ -6.5\% $ & $ +2.2\% $ & $ 0.54 $ & $ +3.7\% $ & $ +1.9\% $ & $ 0.24 $ & $ -4.2\% $ & $ +0.0\% $ & $ 0.62 $ & $ +0.0\% $ & $ +0.0\% $ & $ 0.43 $ & $ +0.0\% $ & $ +0.0\% $ & $ 0.78 $ & \cellcolor{green!12} $ +2.6\% $ &  $ +0.0\% $ & $ 0.66 $ & $ -1.5\% $ & $ -3.0\% $ \\
\cellcolor{politics!50} partisan & $ 0.49 $ & $ +0.0\% $ & $ +2.0\% $ & $ 0.46 $ & $ -2.2\% $ & $ +6.5\% $ & $ 0.54 $ & $ +3.7\% $ & $ +1.9\% $ & $ 0.24 $ & $ -4.2\% $ & $ -4.2\% $ & $ 0.62 $ & $ -1.6\% $ & $ -1.6\% $ & $ 0.43 $ & $ +0.0\% $ & $ -2.3\% $ & $ 0.78 $ & \cellcolor{green!12} $ +2.6\% $ & \cellcolor{green!12} $ +2.6\% $ & $ 0.66 $ & $ -3.0\% $ & $ -3.0\% $ \\
\midrule
\cellcolor{psychology!50} psychologist & $ 0.49 $ & $ +0.0\% $ & $ +0.0\% $ & $ 0.46 $ & $ -2.2\% $ & $ +0.0\% $ & $ 0.54 $ & $ +1.9\% $ & $ +0.0\% $ & $ 0.24 $ & $ -4.2\% $ & $ +0.0\% $ & $ 0.62 $ & $ -1.6\% $ & $ +0.0\% $ & $ 0.43 $ & $ +0.0\% $ & $ +0.0\% $ & $ 0.78 $ & $ +0.0\% $ & $ +0.0\% $ & $ 0.66 $ &  $ +0.0\% $ &  $ +0.0\% $ \\
    \bottomrule
    \end{tabular}
    }}
    \caption{Performance differences (\%) of \textit{activation addition} for Qwen-7B-Chat across roles, relative to the baseline. Positive values indicate performance gains. Highlighted cells show in-domain splits.}
    \label{tab:performance}
    \end{table*}

\paragraph{RQ2: Analysis of directions corresponding to the role.}

For every model \(m \in \mathcal{M}\) and each direction \(d_{i,r}^{(l)}\), we apply the \textit{patch scoping} procedure (detailed in Section~\ref{sec:methods}) to prompt Claude 3.5 Haiku~\cite{anthropic2024haiku}. We ask the model whether the applied direction aligns with the corresponding role \(r \in R\) with the prompt shown in Fig.~\ref{fig:eval_patch_scoping_example}.
Tab.~\ref{tab:performance} shows, among the directions that improve the baseline score on the domain-specific test split \(D_c \in \mathcal{D}_{\text{test}}\), the percentage of those \(d_{i,r}^{(l)}\) that Claude identifies as role-specific.

\begin{figure}[htb]
\centering
\begin{minipage}{0.95\linewidth}
\begin{tcolorbox}[
    title=Prompt to Evaluate Patch Scoping,
    colback=violet!5!white,colframe=violet!75!black,
    width=\textwidth  
]
\small
We are testing whether the LLM has been successfully steered from its baseline output to adopt a \textit{\{role\}} perspective. Your task is to determine if the response contains relevant content to the \textit{\{role\}} domain. Note that you are not assessing the quality of the response—only its relation to the \textit{\{role\}} concept.

\smallskip
Please keep in mind:
(i) The text may be repetitive or somewhat incoherent.
(ii) If the response closely mirrors the baseline without introducing any distinct \textit{\{role\}}-related elements, it should be considered as not aligned with the \textit{\{role\}} role.

Evaluate this step by step and answer the following question:
Is the model being steered toward the \textit{\{role\}} role?

\smallskip
Text to Evaluate: \textit{\{response\_text\}}; Baseline Reference:
\textit{\{baseline\_response\}}.

\end{tcolorbox}
\end{minipage}

\caption{Prompt for evaluating patch scoping output provided to Claude 3.5 Haiku.}
\label{fig:eval_patch_scoping_example}
\end{figure}

\begin{table*}[ht]
\centering
{\small
\resizebox{\textwidth}{!}{
\begin{tabular}{lllllll|l}
\toprule
gemma-2-2b & gemma-2-9b & Llama-3.2-1B & Llama-3.2-3B & Llama-3.1-8B & Qwen-1.8B & Qwen-7B & Overall \\
\midrule
\cellcolor{blue!8} 169/1008 (17\%) & \cellcolor{blue!15} 371/1217 (30\%) & \cellcolor{blue!3} 36/538 (7\%) & \cellcolor{blue!8} 168/987 (17\%) & \cellcolor{blue!7} 87/581 (15\%) & \cellcolor{blue!3} 76/1195 (6\%) & \cellcolor{blue!7} 187/1284 (15\%) & \cellcolor{blue!8} 1094/6810 (16\%) \\
\bottomrule
\end{tabular}
}}
\caption{ Number and percentage of directions interpreted as the corresponding roles by Claude 3.5 Haiku among those that improve upon the baseline, sorted by model family and size.}
\label{tab:percentage}
\end{table*}

\paragraph{RQ3: Directional ablation analysis.}
To evaluate whether the optimal direction \(d^*(\alpha)\) plays a causal role in boosting performance on the test dataset \(\mathcal{D}_{\text{test}}\), we ablate \(d^*(\alpha)\) (with \(\alpha=1\)) in Qwen-7B-Chat and present the resulting performance for each role \(r \in R\) in Tab.~\ref{tab:performance_abl}.

\begin{table*}[ht]
\centering
{\small
\resizebox{\textwidth}{!}{
\begin{tabular}{l|ll|ll|ll|ll|ll|ll|ll|ll}
\toprule
Dataset Split & \multicolumn{2}{r}{\cellcolor{economics!50} Economics (492)} & \multicolumn{2}{r}{\cellcolor{eecs!50} EECS (247)} & \multicolumn{2}{r}{\cellcolor{law!50} Law (200)} & \multicolumn{2}{r}{\cellcolor{math!50} Math (287)} & \multicolumn{2}{r}{\cellcolor{medicine!50} Medicine (241)} & \multicolumn{2}{r}{\cellcolor{natural!50} Natural Science (590)} & \multicolumn{2}{r}{\cellcolor{politics!50} Politics (200)} & \multicolumn{2}{r}{\cellcolor{psychology!50} Psychology (200)} \\
Role $\downarrow \alpha \rightarrow$ & Baseline & Ablation & Baseline & Ablation & Baseline & Ablation & Baseline & Ablation & Baseline & Ablation & Baseline & Ablation & Baseline & Ablation & Baseline & Ablation \\
\midrule
\cellcolor{economics!50} economic researcher & $ 0.49 $ &  $ +0.0\% $ & $ 0.46 $ & $ +4.3\% $ & $ 0.54 $ & $ -3.7\% $ & $ 0.24 $ & $ +0.0\% $ & $ 0.62 $ & $ +1.6\% $ & $ 0.43 $ & $ +2.3\% $ & $ 0.78 $ & $ -3.8\% $ & $ 0.66 $ & $ -3.0\% $ \\
\cellcolor{economics!50} economist & $ 0.49 $ & \cellcolor{red!32} $ -24.5\% $ & $ 0.46 $ & $ -23.9\% $ & $ 0.54 $ & $ -37.0\% $ & $ 0.24 $ & $ -8.3\% $ & $ 0.62 $ & $ -45.2\% $ & $ 0.43 $ & $ -20.9\% $ & $ 0.78 $ & $ -43.6\% $ & $ 0.66 $ & $ -39.4\% $ \\
\cellcolor{economics!50} financial analyst & $ 0.49 $ & \cellcolor{red!40} $ -55.1\% $ & $ 0.46 $ & $ -41.3\% $ & $ 0.54 $ & $ -48.1\% $ & $ 0.24 $ & $ -16.7\% $ & $ 0.62 $ & $ -66.1\% $ & $ 0.43 $ & $ -53.5\% $ & $ 0.78 $ & $ -65.4\% $ & $ 0.66 $ & $ -65.2\% $ \\
\midrule
\cellcolor{eecs!50} electronics technician & $ 0.49 $ & $ -8.2\% $ & $ 0.46 $ & \cellcolor{red!26} $ -13.0\% $ & $ 0.54 $ & $ -3.7\% $ & $ 0.24 $ & $ -8.3\% $ & $ 0.62 $ & $ -6.5\% $ & $ 0.43 $ & $ -11.6\% $ & $ 0.78 $ & $ -10.3\% $ & $ 0.66 $ & $ -16.7\% $ \\
\cellcolor{eecs!50} data scientist & $ 0.49 $ & $ -44.9\% $ & $ 0.46 $ & \cellcolor{red!35} $ -34.8\% $ & $ 0.54 $ & $ -29.6\% $ & $ 0.24 $ & $ -20.8\% $ & $ 0.62 $ & $ -56.5\% $ & $ 0.43 $ & $ -48.8\% $ & $ 0.78 $ & $ -42.3\% $ & $ 0.66 $ & $ -43.9\% $ \\
\cellcolor{eecs!50} electrical engineer & $ 0.49 $ & $ -2.0\% $ & $ 0.46 $ & \cellcolor{red!16} $ -4.3\% $ & $ 0.54 $ & $ +0.0\% $ & $ 0.24 $ & $ -4.2\% $ & $ 0.62 $ & $ +1.6\% $ & $ 0.43 $ & $ -11.6\% $ & $ 0.78 $ & $ -7.7\% $ & $ 0.66 $ & $ -10.6\% $ \\
\cellcolor{eecs!50} software engineer & $ 0.49 $ & $ +2.0\% $ & $ 0.46 $ & \cellcolor{red!16} $ -4.3\% $ & $ 0.54 $ & $ -3.7\% $ & $ 0.24 $ & $ -8.3\% $ & $ 0.62 $ & $ -1.6\% $ & $ 0.43 $ & $ -2.3\% $ & $ 0.78 $ & $ -5.1\% $ & $ 0.66 $ & $ -3.0\% $ \\
\cellcolor{eecs!50} web developer & $ 0.49 $ & $ -42.9\% $ & $ 0.46 $ & \cellcolor{red!36} $ -37.0\% $ & $ 0.54 $ & $ -24.1\% $ & $ 0.24 $ & $ -25.0\% $ & $ 0.62 $ & $ -40.3\% $ & $ 0.43 $ & $ -44.2\% $ & $ 0.78 $ & $ -26.9\% $ & $ 0.66 $ & $ -30.3\% $ \\
\midrule
\cellcolor{law!50} bailiff & $ 0.49 $ & $ -2.0\% $ & $ 0.46 $ & $ -2.2\% $ & $ 0.54 $ & \cellcolor{red!10} $ -1.9\% $ & $ 0.24 $ & $ -4.2\% $ & $ 0.62 $ & $ +1.6\% $ & $ 0.43 $ & $ +2.3\% $ & $ 0.78 $ & $ -1.3\% $ & $ 0.66 $ & $ -3.0\% $ \\
\cellcolor{law!50} lawyer & $ 0.49 $ & $ -6.1\% $ & $ 0.46 $ & $ -2.2\% $ & $ 0.54 $ & \cellcolor{red!15} $ -3.7\% $ & $ 0.24 $ & $ -4.2\% $ & $ 0.62 $ & $ -1.6\% $ & $ 0.43 $ & $ -9.3\% $ & $ 0.78 $ & $ -3.8\% $ & $ 0.66 $ & $ -6.1\% $ \\
\midrule
\cellcolor{math!50} data analyst & $ 0.49 $ & $ -32.7\% $ & $ 0.46 $ & $ -26.1\% $ & $ 0.54 $ & $ -27.8\% $ & $ 0.24 $ & \cellcolor{red!22} $ -8.3\% $ & $ 0.62 $ & $ -35.5\% $ & $ 0.43 $ & $ -25.6\% $ & $ 0.78 $ & $ -29.5\% $ & $ 0.66 $ & $ -36.4\% $ \\
\cellcolor{math!50} mathematician & $ 0.49 $ & $ +0.0\% $ & $ 0.46 $ & $ +2.2\% $ & $ 0.54 $ & $ +0.0\% $ & $ 0.24 $ & \cellcolor{red!16} $ -4.2\% $ & $ 0.62 $ & $ +0.0\% $ & $ 0.43 $ & $ +0.0\% $ & $ 0.78 $ & $ -1.3\% $ & $ 0.66 $ & $ -3.0\% $ \\
\cellcolor{math!50} statistician & $ 0.49 $ & $ -55.1\% $ & $ 0.46 $ & $ -41.3\% $ & $ 0.54 $ & $ -55.6\% $ & $ 0.24 $ & \cellcolor{red!32} $ -25.0\% $ & $ 0.62 $ & $ -67.7\% $ & $ 0.43 $ & $ -58.1\% $ & $ 0.78 $ & $ -66.7\% $ & $ 0.66 $ & $ -69.7\% $ \\
\midrule
\cellcolor{medicine!50} nurse & $ 0.49 $ & $ +0.0\% $ & $ 0.46 $ & $ +2.2\% $ & $ 0.54 $ & $ -9.3\% $ & $ 0.24 $ & $ +8.3\% $ & $ 0.62 $ & \cellcolor{red!22} $ -8.1\% $ & $ 0.43 $ & $ -4.7\% $ & $ 0.78 $ & $ -9.0\% $ & $ 0.66 $ & $ -12.1\% $ \\
\cellcolor{medicine!50} doctor & $ 0.49 $ & $ -18.4\% $ & $ 0.46 $ & $ -8.7\% $ & $ 0.54 $ & $ -22.2\% $ & $ 0.24 $ & $ -4.2\% $ & $ 0.62 $ & \cellcolor{red!32} $ -25.8\% $ & $ 0.43 $ & $ -14.0\% $ & $ 0.78 $ & $ -21.8\% $ & $ 0.66 $ & $ -19.7\% $ \\
\cellcolor{medicine!50} physician & $ 0.49 $ & $ +0.0\% $ & $ 0.46 $ & $ +2.2\% $ & $ 0.54 $ & $ +0.0\% $ & $ 0.24 $ & $ +0.0\% $ & $ 0.62 $ & \cellcolor{red!14} $ -3.2\% $ & $ 0.43 $ & $ +0.0\% $ & $ 0.78 $ & $ -1.3\% $ & $ 0.66 $ & $ -3.0\% $ \\
\cellcolor{medicine!50} dentist & $ 0.49 $ & $ -53.1\% $ & $ 0.46 $ & $ -37.0\% $ & $ 0.54 $ & $ -50.0\% $ & $ 0.24 $ & $ +0.0\% $ & $ 0.62 $ & \cellcolor{red!41} $ -59.7\% $ & $ 0.43 $ & $ -41.9\% $ & $ 0.78 $ & $ -66.7\% $ & $ 0.66 $ & $ -59.1\% $ \\
\cellcolor{medicine!50} surgeon & $ 0.49 $ & $ -8.2\% $ & $ 0.46 $ & $ -13.0\% $ & $ 0.54 $ & $ -9.3\% $ & $ 0.24 $ & $ -12.5\% $ & $ 0.62 $ & \cellcolor{red!26} $ -12.9\% $ & $ 0.43 $ & $ -14.0\% $ & $ 0.78 $ & $ -12.8\% $ & $ 0.66 $ & $ -9.1\% $ \\
\midrule
\cellcolor{natural!50} geneticist & $ 0.49 $ & $ -4.1\% $ & $ 0.46 $ & $ +6.5\% $ & $ 0.54 $ & $ +1.9\% $ & $ 0.24 $ & $ +4.2\% $ & $ 0.62 $ & $ -6.5\% $ & $ 0.43 $ & \cellcolor{red!20} $ -7.0\% $ & $ 0.78 $ & $ -3.8\% $ & $ 0.66 $ & $ -10.6\% $ \\
\cellcolor{natural!50} biologist & $ 0.49 $ & $ -8.2\% $ & $ 0.46 $ & $ +6.5\% $ & $ 0.54 $ & $ +0.0\% $ & $ 0.24 $ & $ +8.3\% $ & $ 0.62 $ & $ -11.3\% $ & $ 0.43 $ & \cellcolor{red!23} $ -9.3\% $ & $ 0.78 $ & $ -5.1\% $ & $ 0.66 $ & $ -10.6\% $ \\
\cellcolor{natural!50} physicist & $ 0.49 $ & $ +0.0\% $ & $ 0.46 $ & $ -2.2\% $ & $ 0.54 $ & $ -1.9\% $ & $ 0.24 $ & $ -4.2\% $ & $ 0.62 $ & $ -1.6\% $ & $ 0.43 $ &  $ +0.0\% $ & $ 0.78 $ & $ +0.0\% $ & $ 0.66 $ & $ -4.5\% $ \\
\cellcolor{natural!50} teacher & $ 0.49 $ & $ -57.1\% $ & $ 0.46 $ & $ -41.3\% $ & $ 0.54 $ & $ -44.4\% $ & $ 0.24 $ & $ -25.0\% $ & $ 0.62 $ & $ -64.5\% $ & $ 0.43 $ & \cellcolor{red!39} $ -51.2\% $ & $ 0.78 $ & $ -59.0\% $ & $ 0.66 $ & $ -66.7\% $ \\
\cellcolor{natural!50} chemist & $ 0.49 $ & $ -2.0\% $ & $ 0.46 $ & $ +6.5\% $ & $ 0.54 $ & $ -3.7\% $ & $ 0.24 $ & $ +0.0\% $ & $ 0.62 $ & $ +0.0\% $ & $ 0.43 $ & \cellcolor{red!11} $ -2.3\% $ & $ 0.78 $ & $ +0.0\% $ & $ 0.66 $ & $ -4.5\% $ \\
\cellcolor{natural!50} ecologist & $ 0.49 $ & $ +0.0\% $ & $ 0.46 $ & $ -2.2\% $ & $ 0.54 $ & $ +1.9\% $ & $ 0.24 $ & $ +0.0\% $ & $ 0.62 $ & $ +1.6\% $ & $ 0.43 $ &  $ +0.0\% $ & $ 0.78 $ & $ -1.3\% $ & $ 0.66 $ & $ -1.5\% $ \\
\midrule
\cellcolor{politics!50} politician & $ 0.49 $ & $ +0.0\% $ & $ 0.46 $ & $ +4.3\% $ & $ 0.54 $ & $ +0.0\% $ & $ 0.24 $ & $ +4.2\% $ & $ 0.62 $ & $ +1.6\% $ & $ 0.43 $ & $ +0.0\% $ & $ 0.78 $ & \cellcolor{red!12} $ -2.6\% $ & $ 0.66 $ & $ -4.5\% $ \\
\cellcolor{politics!50} sheriff & $ 0.49 $ & $ -2.0\% $ & $ 0.46 $ & $ +0.0\% $ & $ 0.54 $ & $ -1.9\% $ & $ 0.24 $ & $ -4.2\% $ & $ 0.62 $ & $ +1.6\% $ & $ 0.43 $ & $ +2.3\% $ & $ 0.78 $ & \cellcolor{red!12} $ -2.6\% $ & $ 0.66 $ & $ -3.0\% $ \\
\cellcolor{politics!50} enthusiast & $ 0.49 $ & $ -4.1\% $ & $ 0.46 $ & $ +2.2\% $ & $ 0.54 $ & $ +0.0\% $ & $ 0.24 $ & $ -4.2\% $ & $ 0.62 $ & $ -1.6\% $ & $ 0.43 $ & $ +0.0\% $ & $ 0.78 $ & \cellcolor{red!12} $ -2.6\% $ & $ 0.66 $ & $ -3.0\% $ \\
\cellcolor{politics!50} partisan & $ 0.49 $ & $ +0.0\% $ & $ 0.46 $ & $ +2.2\% $ & $ 0.54 $ & $ -1.9\% $ & $ 0.24 $ & $ -8.3\% $ & $ 0.62 $ & $ -1.6\% $ & $ 0.43 $ & $ +2.3\% $ & $ 0.78 $ & \cellcolor{red!12} $ -2.6\% $ & $ 0.66 $ & $ -3.0\% $ \\
\midrule
\cellcolor{psychology!50} psychologist & $ 0.49 $ & $ -2.0\% $ & $ 0.46 $ & $ -2.2\% $ & $ 0.54 $ & $ -3.7\% $ & $ 0.24 $ & $ +0.0\% $ & $ 0.62 $ & $ -1.6\% $ & $ 0.43 $ & $ +0.0\% $ & $ 0.78 $ & $ -1.3\% $ & $ 0.66 $ & \cellcolor{red!10} $ -1.5\% $ \\
\bottomrule
\end{tabular}
}}
\caption{Performance differences (\%) of  of \textit{directional ablation} across roles for Qwen-7B-Chat, relative to the baseline. Negative values indicate expected performance drops. Highlighted cells show in-domain splits.}
\label{tab:performance_abl}
\end{table*}

\section{Discussion}
\label{sec:discussion}

\paragraph{The effect of steering on model performances.}
In larger models, performance generally increases in the target and related domains and worsens or remains unchanged in domains unrelated to the role. Since we observe a strong correlation among larger models, as shown in Fig.~\ref{fig:corrmatrix}, we show the full details of one model, Qwen-7B, in Tab.~\ref{tab:performance} due to space constraints\footnote{Full experimental results are available to reviewers in supplementary materials.}. For instance, we notice that the mathematician's role significantly improves Qwen-7B's performance in the domain test set, math, and the related field of EECS. Similarly, the doctor's role improves primarily medicine but also natural science. On the other hand, the direction corresponding to the psychologist's role does not yield any performance benefit in its designated reference split. Furthermore, we did not identify any alternative steering directions that would enhance performance in psychology.

Our analysis in Tab.~\ref{tab:performance} reveals that when evaluating the optimal steering direction, applying an \textit{activation addition} with a coefficient of \(\alpha = 3.0\) results in performance that exceeds or is comparable to \(\alpha = 1.0\) and the baseline model. This suggests that a higher intervention intensity may more effectively align the model’s internal representations with the desired domain-specific features, enhancing its performance on targeted tasks.
In smaller models, on the other hand, performance increases both in-domain and out-of-domain.

While~\cite{zheng2024helpful} finds that adding a role through prompting can lead to unpredictable performance gains; we show that modifying internal representations more precisely steers the LLM to perform better on target domain tasks.

\paragraph{Are directions capturing the role?}
We observe that role-based interventions often produce directional shifts in the model’s activation space that enhance performance within the target domain and, in some cases (e.g., as evidenced by \(d^*(\alpha)\) in Tab.~\ref{tab:performance}), in closely related domains. However, these directions are not always directly interpretable and do not correspond to the intended roles. As shown in Tab.~\ref{tab:percentage}, on average, only 16\% of the directions yielding improvements in the relevant test split are directly interpreted by Claude 3.5 Haiku as reflecting the intended role. In other words, while some of the identified activation directions benefit performance, they do not necessarily align with the semantic role as determined by patch-scoping methods. This is a known characteristic of patch-scoping~\cite{kharlapenko2024self} that distinguishes it from auto-interp~\cite{bills2023language}. While auto-interp leverages the feature's maximally activating examples from the training set of SAEs to prompt a language model to interpret that feature, patch-scoping captures the underlying concept represented by the feature yet struggles to explicitly identify the "label" of the concept.

Examining Tab.~\ref{tab:percentage}, we notice that larger models exhibit activation directions more clearly interpretable as corresponding to specific roles than their smaller counterparts within a given model family. This observation holds for Gemma-2, Qwen, and Llama-3.2\footnote{Note that a direct comparison between Llama-3.1 and Llama-3.2 is not feasible, as their pre-training and post-training methodologies differ.}.
This indicates larger models can capture and encode fine-grained role-specific features within their activation spaces. In contrast, smaller models tend to develop more general, abstract representations that may blend multiple role-related cues, making it harder to isolate a clear directional signal corresponding to a specific role. This aligns with the evidence from Anthropic in~\cite{templeton2024scaling} that as model scale increases, representations become more mono-semantic, meaning activations align more closely with specific concepts.

\paragraph{The effect of ablating roles.}
Results in Tab.~\ref{tab:performance_abl} indicate that ablating the optimal \(d^*(\alpha)\) activation directions yields heterogeneous effects. 
For directions associated with the role \( r \in R \) that correspond to the domain-specific dataset \( D_c \in \mathcal{D}_\textit{test} \), where \( c \sim r \), we observe a performance degradation, which aligns with expectations. Notably, in domain-specific datasets \( D_c \in \mathcal{D}_\textit{test} \) unrelated to the role \( r \), where \( c \not\sim r \), performance generally declines but occasionally exhibits a marginal improvement. We hypothesise that this variation arises because the removal process may eliminate certain noise components without significantly disrupting the core representational structure essential for the task. As shown by~\cite{dalvi2020analyzing}, many neurons across neural networks are redundant and can be removed when optimising towards a downstream task.
Also, Tab.~\ref{tab:percentage} clearly shows that multiple directions exist in the activation space that yields an improvement; ablating a single direction can remove noise and amplify the effect of the remaining ones. 
Smaller models have less redundancy, making them more sensitive to perturbations. While steering interventions can enhance performance, they can just as easily cause deterioration across test splits. This sensitivity likely stems from more concentrated representations, where each directional component is crucial for encoding domain-specific knowledge.



\section{Conclusion}
\label{sec:conclusion}

In this work, we introduced role vectors as a novel method for guiding the behaviour of LLMs by directly manipulating their internal activations. By computing difference‐in‐means vectors between role-specific prompts and a generic baseline, our approach shows that targeted activation addition can steer models toward domain-specific expertise. Our experiments, spanning multiple models and diverse domains, reveal that such interventions can enhance task performance in the target domain while largely preserving general capabilities. We also show that the effectiveness of role-based steering is sensitive to both model scale and the depth at which the intervention is applied; larger models and deeper layers tend to yield more robust and interpretable directional signals.

Notably, our patch-scoping analysis indicates that only a subset of the activation directions aligns with the intended roles, underscoring the complexity of internal model representations.
Future work will study this mechanistically using Activation Patching (Causal Mediation Analysis) techniques using SAE features~\cite{heimersheim2024use} to explain this phenomenon better.

Our findings suggest that embedding role vectors within model activations offers a promising pathway for achieving more controllable behaviour in large language models. Our further work will explore this phenomenon in greater depth, considering additional analytical dimensions and potential biases.

\section*{Acknowledgments}

The experiments of this work have been conducted on Leonardo supercomputer with the support of CINECA-Italian Super Computing Resource Allocation, class C project IsCb7\_LLM-EVAL (HP10CIO7T9).

\section*{Limitations}


Although we examined a diverse set of open-source models, our results might differ in untested models, especially larger ones.
Additionally, our analysis does not offer a complete mechanistic explanation of the phenomenon, a different methodology that will be explored in future research. While we pinpointed a specific direction influencing performances in each model, its exact semantic interpretation remains uncertain. The term “role direction” is used functionally here, but these directions might represent other underlying concepts.
While the targeted domain performance improves, applying role vectors might degrade performance in unrelated tasks, making the intervention less universally beneficial. Steering models using role vectors may inadvertently reinforce biases or lead to overconfidence in certain domains. Careful evaluation will be conducted in future works to mitigate unintended consequences. 

\bibliography{bibliography}



\end{document}